\documentclass{article}
\usepackage{amsmath,amsfonts,amssymb}
\usepackage{algorithmic}
\usepackage{algorithm}
\usepackage{array}
\usepackage{textcomp}
\usepackage{stfloats}
\usepackage{url}
\usepackage{verbatim}
\usepackage{graphicx}
\usepackage{mathtools}
\usepackage{bm}
\usepackage{hyperref}
\usepackage{nicefrac}
\usepackage{booktabs}
\usepackage{subcaption}
\usepackage{epsfig}
\usepackage{spconf}
\usepackage{xcolor}
\usepackage{fancyhdr}

\let\OLDthebibliography\thebibliography
\renewcommand\thebibliography[1]{
  \OLDthebibliography{#1}
  \setlength{\parskip}{0pt}
  \setlength{\itemsep}{0pt plus 0.3ex}
}

\fancypagestyle{copyright}{
    \fancyhf{}
    \fancyhead[C]{
        \small
        \textit{IEEE ICME Workshop on Coding for Machines, Niagara Falls, Canada, 2024.} \\
        \vspace{10pt}
        © 2024 IEEE. Personal use of this material is permitted. Permission from IEEE must be obtained for all other uses, in any current or future media, including reprinting/republishing this material for advertising or promotional purposes, creating new collective works, for resale or redistribution to servers or lists, or reuse of any copyrighted component of this work in other works.
    }
    
}

\DeclareMathOperator*{\argmax}{arg\,max}

\DeclareMathSymbol{\shortminus}{\mathbin}{AMSa}{"39}

\begin{document}\sloppy
\setlength{\abovedisplayskip}{4pt}
\setlength{\belowdisplayskip}{4pt}

\title{Towards Task-Compatible Compressible Representations}
%
\name{Anderson de Andrade and Ivan V. Baji\'{c}}
\address{School of Engineering Science, Simon Fraser University, Burnaby, BC, Canada \\\normalsize{\texttt{anderson\_de\_andrade@sfu.ca}, \texttt{ibajic@ensc.sfu.ca}}
}

\maketitle
\thispagestyle{copyright}

\begin{abstract}
We identify an issue in multi-task learnable compression, in which a representation learned for one task does not positively contribute to the rate-distortion performance of a different task as much as expected, given the estimated amount of information available in it. We interpret this issue using the predictive $\mathcal{V}$-information framework. In learnable scalable coding, previous work increased the utilization of side-information for input reconstruction by also rewarding input reconstruction when learning this shared representation. We evaluate the impact of this idea in the context of input reconstruction more rigorously and extended it to other computer vision tasks. We perform experiments using representations trained for object detection on COCO 2017 and depth estimation on the Cityscapes dataset, and use them to assist in image reconstruction and semantic segmentation tasks. The results show considerable improvements in the rate-distortion performance of the assisted tasks. Moreover, using the proposed representations, the performance of the base tasks are also improved. Results suggest that the proposed method induces simpler representations that are more compatible with downstream processes.
\end{abstract}


\section{Introduction}
Learnable compression methods often rely on the information bottleneck framework \cite{tishby2000information}, in which a constrained probability distribution is enforced on an intermediate representation as it is fitted to perform a predictive task. The resulting probability model can be used to efficiently code this representation. In scalable coding for humans and machines \cite{Choi2022ScalableIC}, a compressible representation from a computer vision task is used as side information to assist in a reconstruction task. As an efficient representation is learned for a particular task, there is the possibility that the relevant information present in it cannot be fully extracted by a different process. This phenomenon is usually referred to as co-adaptation in the transfer learning literature \cite{YosinskiCBL14, Ben-DavidBCKPV10}. From an information theory point of view, the framework of predictive $\mathcal{V}$-information \cite{XuZSSE20} can also be used to interpret this issue more formally.

In this work, we 
evaluate the ability of the technique proposed by \cite{AndradeHFB23} to induce representations that are more compatible with 
secondary tasks. By adding an auxiliary reconstruction task to the rate-distortion formulation, we create representations that not only outperform in rate-distortion performance of secondary tasks, but also have better rate-distortion performance on the 
base task. We perform our analysis in the context of learnable scalable coding, where a task uses a learned representation for a different task as side-information in addition to its own representation. Current approaches for scalable coding \cite{Choi2022ScalableIC, ForoutanHAB23} are limited in the amount of information re-utilized from their shared representations. We provide theoretical understanding and perform experiments to show that comparable rate-distortion performance can be achieved on 
base tasks while the resulting representations still perform effectively on 
secondary tasks.

\section{Related Work}
Learnable lossy compression models \cite{BalleLS17, balle2018variational, Cheng2020LearnedIC, HeYPMQW22} are often formulated as variational autoencoders (VAEs) \cite{Kingma2014AutoEncodingVB}, in which a parameterized probability distribution is fitted to the latent representation. As per the information bottleneck framework~\cite{tishby2000information}, the entropy of this probability distribution is constrained by a Lagrange multiplier, and the system is optimized to both fit the latent representation to a probability distribution and minimize the task performance using this representation.

To effectively encode these latent representations, the probability model is designed to infer on a group of elements conditioned on a subset of previously observed elements. The latent representation is then quantized and encoded using the probability model to produce a bitstream at a resulting rate. In these methods, the loss of information arises from the quantization of the latent representation and from the analysis transform, as shown by the data processing inequality \cite{0016881}.

Entropy models \cite{balle2018variational, HeYPMQW22} are developed to be used in conjunction with a lossy encoder \cite{JiangYZNGW23, MLIC++}. Their architectures usually divide the dimensions of the latent representation into groups for parallel processing. Most of these models can use side-information as context for conditional coding. We make use of this feature to perform experiments in which we test the performance of our proposed methods using the scalable conditional coding from~\cite{AndradeHFB23}. While~\cite{AndradeHFB23} focused on conditional and residual approaches for coding the enhancement layer for input reconstruction, in this paper our focus is on compatibility of base representation with secondary tasks, which may or may not be input reconstruction. The scalable coding framework from~\cite{AndradeHFB23} is used as the platform for this study.

Most work on scalable learnable coding \cite{Choi2022ScalableIC} considers two tasks: a computer vision task (base task) and the image reconstruction task (enhancement task). It is often assumed that the information available in the latent representation of the computer vision task is a subset of the information required for image reconstruction. As such, different methods have been proposed to complement the information available in the latent representation of the computer vision task for the reconstruction task. To increase the amount of utilization of this base representation, the work in~\cite{AndradeHFB23} proposed to train the latent representation to reconstruct the input. In this work, we provide theoretical analysis, extend this idea to non-reconstruction tasks, and evaluate this approach.

Predictive $\mathcal{V}$-information \cite{XuZSSE20} extends the concepts of Shannon's information theory to consider the computational limitations. 
These concepts explain situations in which computation can increase the amount of information perceived, which would seem to violate the classical data processing inequality~\cite{0016881}. We use them to demonstrate how our representations can be compatible with other task models designed to work on the original input space.

\section{Proposed Method}

\begin{figure*}
  \centering
  \includegraphics[width=\textwidth]{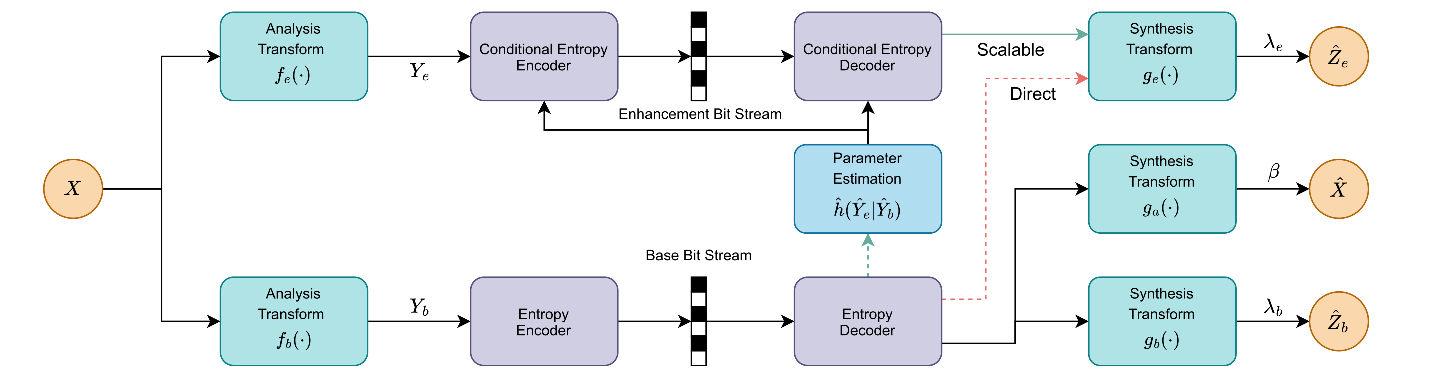}
  \caption{Architecture overview. The green lines denote the scalable approach, and the red line denotes a direct approach where the secondary task only has access to the base representation. The dotted lines mean that  gradients do not flow past that edge.}
  \label{figure:diagram}
\end{figure*}

We define the problem in terms of predictive $\mathcal{V}$-information and 
reframe the solution presented in \cite{AndradeHFB23} accordingly. 
After a brief theoretical analysis, we describe our model architecture and define the scalable learnable compression setup.

\subsection{Problem Formulation}
A learnable compressor for input $X$ and target $Z$ defines the Markov chain $X \to Y \to \hat{Z}$, where $\hat{Z}$ tries to approximate $Z$. Let $\mathcal{F}$ be the set of all analysis transforms such that $\mathbf{f} \in \mathcal{F}; \mathbf{f}:\mathcal{X} \to \mathcal{Y}; \mathcal{Y} \subseteq \mathbb{Z}^M; M \in \mathbb{N}^+.$ 
Also, let $\mathcal{G}$ be the set of all \textit{attainable}\footnote{Given the architecture, complexity constraints, etc.} synthesis transforms, with $\mathbf{g} \in \mathcal{G}; \mathbf{g} : \mathcal{Y} \to \mathcal{Z}$. Using the $\mathcal{V}$-information \cite{XuZSSE20} framework, the objective of the learnable compressor can be expressed as:
\begin{align*}
\sup_{\mathbf{f} \in \mathcal{F}} I_{\mathcal{V}}(\mathbf{f}(X) \to Z)
&=
\sup_{\mathbf{f} \in \mathcal{F}} 
\left[
H_{\mathcal{V}}(Z|\emptyset) - H_{\mathcal{V}}(Z|\mathbf{f}(X))
\right]
,
\end{align*}
where $I_\mathcal{V}$ is the predictive $\mathcal{V}$-information and $H_{\mathcal{V}}(\cdot|\cdot)$ is the predictive conditional $\mathcal{V}$-entropy, defined as:
\begin{align*}
    H_{\mathcal{V}}(Z|Y)
    =
    \inf_{h \in \mathcal{V}}
    \mathbb{E}_{\mathbf{z}, \mathbf{y} \sim Z,Y}
    \left[
    - \log h[\mathbf{y}](\mathbf{z})
    \right]
    ,
\end{align*}
where:
\begin{align*}
    \mathcal{V} \subset \{h : \mathbf{y} \to \mathcal{P}(\mathcal{Z}; \mathbf{g}(\mathbf{y})), \mathbf{y} \in \mathcal{Y} ; \emptyset \to P(\mathcal{Z}) | \mathbf{g} \in \mathcal{G} \}
\end{align*}
is a set of predictive models that can be attained by the choice of model and optimization algorithm. $\mathcal{P}(\mathcal{Z}; \mathbf{g}(\mathbf{y}))$ denotes the set of all probability measures 
on $\mathcal{Z}$, parameterized by $\mathbf{g}(\mathbf{y})$. Thus, in this analysis, attaining the mutual information $I(X;Z)$ is limited by the expressiveness of $\mathcal{G}$ and $\mathcal{V}$, but not $\mathcal{F}$, since it includes all analysis transforms.

\subsection{Task-Compatible Representations}

Let $\mathbf{f}^*_b = \argmax_{\mathbf{f} \in \mathcal{F}} I_{\mathcal{V}_b}(\mathbf{f}(X) \to Z_b)$ be the best analysis transform for the base task $Z_b$ given $\mathcal{V}_b$, and let:
\begin{align*}
\mathcal{F}^*_b
=
\left\{
\mathbf{f} \in \mathcal{F}
:
I_{\mathcal{V}_b}(\mathbf{f}(X) \to Z_b) = I_{\mathcal{V}_b}(\mathbf{f}^*_b(X) \to Z_b)
\right\}
\end{align*}
be the set of functions that achieve the best predictive $\mathcal{V}$-information for that task. For 
the secondary task $Z_e$ and a set of attainable synthesis transforms $\mathcal{G}_e$ with a corresponding predictive family $\mathcal{V}_e$, we have that:
\begin{align}
    I_{\mathcal{V}_e}(\mathbf{f}^*_b(X) \to Z_e)
    \leq
    \argmax_{\mathbf{f} \in \mathcal{F}^*_b} I_{\mathcal{V}_e}(\mathbf{f}(X) \to Z_e)
    .
\end{align}
Thus, it would be ideal to propose a different analysis transform $\hat{\mathbf{f}}_b \in \mathcal{F}^*_b$ that achieves a higher predictive $\mathcal{V}$-information on task $Z_e$ while retaining the same performance on task $Z_b$. To induce this, we propose the following objective:
\begin{align}
    \sup_{\mathbf{f} \in \mathcal{F}}
    \left\{
    I_{\mathcal{V}_b}(\mathbf{f}(X) \to Z_b)
    +
    \beta
    I_{\mathcal{V}_a}(\mathbf{f}(X) \to Z_a)
    \right\}
    ,
    \label{eq:base_V-info}
\end{align}
where $Z_a$ is an auxiliary task, $\beta$ is a sufficiently small positive scalar, and $\mathcal{V}_a$ is a predictive set given by the set of synthesis transforms $\mathcal{G}_a$. By setting $Z_a = X$, we add a small reward on the capacity of $\mathbf{f}$ to reconstruct the input using a simple transform, preventing the co-adaptation or specialization of $Y$ to the main task $Z_b$. The choice of reconstruction as a task and a simple synthesis transform ensures that an approximation of the original input can be recovered by other tasks. This seems 
to be a good general choice, since task models are often designed to work on the original input space and can be 
trained in an unsupervised manner. We design the architecture defining $\mathcal{G}_a$ to be subsumed by the synthesis transforms $\mathcal{G}_e$ of the secondary task, ensuring that the model has enough expressive power to reconstruct the input and perform the task, if necessary.

A small $\beta$ must be chosen to avoid driving the new objective too far away from the original performance on the primary task. However, considering that the original objective is already a combination of tasks, namely the task objective and the entropy minimization objective, it could be beneficial to use the proposed objective in single-task scenarios. The proposed objective could then prevent the co-adaptation between the representation and the entropy or task models.

\subsection{Model Architecture}

As analysis and synthesis transforms for the vision problem space, we 
use convolutional neural networks (CNN), consisting of scaling operations interleaved with stacks of residual bottleneck blocks \cite{HeZRS16}. The structure of such blocks is the same as \cite{HeYPMQW22} but compared to the overall architecture, there are no attention blocks. As we scale down the spatial resolution we increase the number of channels in the analysis transform and do the inverse for the synthesis transform using transposed convolutions. We have 4 scaling operations, each by a factor of 2. We use Exponential Linear Units (ELU) as activation functions. To perform a concrete task, we connect a task-appropriate neural network to the synthesis transform. 

As an entropy model, we rely on the auto-regressive model of \cite{balle2018variational}. The model leverages masked convolutions to enforce the necessary context constraints. At each step, it processes all channels for a particular spatial dimension. The context corresponds to all dimensions of the previous spatial locations within the receptive field of the CNN. The learnable compression model of \cite{balle2018variational} generates a hyper-prior that is used as side-information by the entropy model.  We use this mechanism to inject the side-information from the primary task, as to conditionally code the representation of the secondary task. The representation of the primary task is processed by a ResNet and concatenated with the context to predict the means and standard deviations of a multivariate normal distribution assigned to the secondary representation. The primary task is encoded without any hyper-prior or side-information.

Fig.~\ref{figure:diagram} shows a diagram of this architecture.\footnote{\href{https://github.com/adeandrade/research}{https://github.com/adeandrade/research}} A direct approach to perform the secondary task is included in which we train the task using the shared representations as input, without a dedicated channel. This architecture will be used to further evaluate the proposed method. 

\subsection{Coding Fidelity}
The quantization of the latent representation during training is simulated in the proposed model of \cite{balle2018variational} using additive uniform noise $\mathcal{U}(\nicefrac{1}{2}, \nicefrac{1}{2})$. This technique is used to allow gradients to back-propagate through the representation. However, it produces small discrepancies between the decoded and original representations, which the synthesis transform must learn to ignore. In some settings, these differences penalize the performance of the task during test time. To avoid this situation, we instead replace the round operations with quantization and straight-through gradients \cite{TheisSCH17} where quantization gradient is set to the identity function. The symbols to be encoded are then computed as $\lfloor \hat{y} - \mu \rceil; \hat{y} = \lfloor y \rceil$, and the decoding process performs $\lfloor \lfloor \hat{y} - \mu \rceil + \mu \rceil$, which recovers $\hat{y}$. We use $\hat{y}$ as input to the entropy model, synthesis transforms, and the rate function, during both training and test time. This ensures that the entropy model produces the same results when using the original representation or generating it auto-regressively. Accurate rate measurements can then be performed without encoding the representation.

\subsection{Scalable Compression}
We employ the information bottleneck \cite{tishby2000information} to produce intermediate lossy representations $Y_b$ and $Y_e$ for tasks $Z_b$ and $Z_e$ respectively. In scalable compression, we refer to the shared representation $Y_b$ as belonging to the \textit{base} task and the dedicated representation $Y_e$ as the \textit{enhancement} task. As presented in \cite{AndradeHFB23}, we add a small reconstruction reward to the rate-distortion Lagrange minimization formulation of the base loss, such that:
\begin{align}
    \nonumber
    &
    \mathcal{L}_b(\psi_b, \phi_b, \theta_b, \theta_a)
    =
    \lambda_b \mathbb{E}[d_b(\hat{Z}_b, Z_b)]
    +
    R_b
    +
    \beta \mathbb{E}[\Vert{ \hat{X} \shortminus X \Vert}]
    ;
    \\
    \nonumber
    &
    Y_b = \mathbf{f}_b(X; \psi_b)
    ,
    \hat{Z}_b = \mathbf{g}_b(\hat{Y}_b; \theta_b)
    ,
    \hat{X} = \mathbf{g}_a(\hat{Y}_b; \theta_a)
    ,
    \\
    &
    R_b = \hat{h}(\hat{Y}_b;\phi_b)
    ,
    \label{eq:base_RD}
\end{align}
where $\hat{h}(\cdot;\phi_b)$ is an entropy estimate parameterized by $\phi_b$, $d_b(\cdot,\cdot)$ is the base task loss. $\lambda_b$ and $\beta$ are hyper-parameters. The $\lambda$ hyper-parameter controls the rate-distortion trade-off. For the enhancement task, we use the traditional rate-distortion loss function:
\begin{align}
    \nonumber
    &
    \mathcal{L}_e(\psi_e, \phi_e)
    =
    \lambda_e \mathbb{E}[d_e(\hat{Z}_e, Z_e)] + \hat{h}(\hat{Y}_e | \hat{Y}_b; \phi_e)
    ;
    \\
    &
    \hat{Y}_e
    =
    \mathbf{f}_e(X;\psi_e)
    ,
    \hat{Z}_e
    =
    \mathbf{g}_e(\hat{Y}_e;\theta_e)
    ,
\end{align}
where $\hat{h}(\cdot|\cdot;\phi_e)$ is the conditional entropy estimate, parameterized by $\phi_e$. During training, either $\mathbf{f}_b(\cdot;\psi_b)$ remains frozen, or the gradients from the enhancement loss do not flow to it.

\section{Experiments}

\begin{figure*}[!ht]
  \captionsetup[subfigure]{aboveskip=0pt}
  \centering
  \begin{subfigure}{0.495\textwidth}
    \centering
    \includegraphics[width=.85\textwidth]{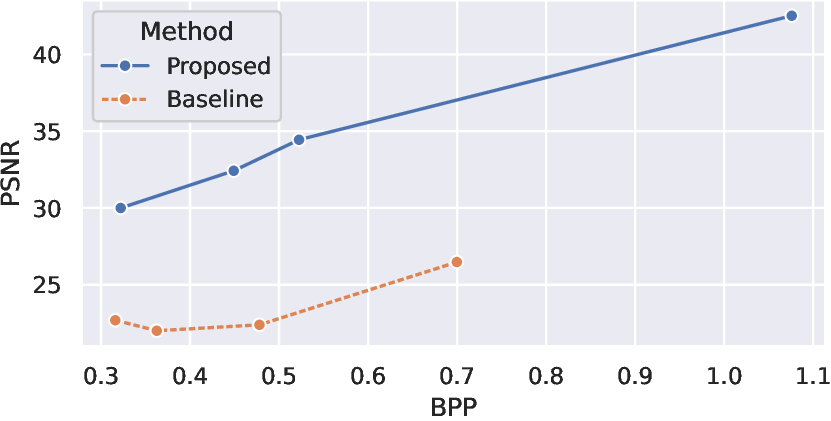}
    \caption{Direct with object detection}
    \label{fig-2a}
  \end{subfigure}
  \begin{subfigure}{0.495\textwidth}
    \centering
    \includegraphics[width=.85\textwidth]{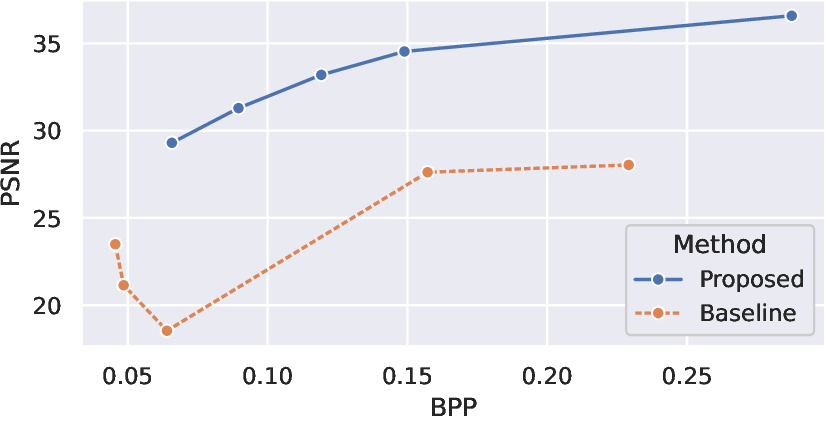}
    \caption{Direct with depth estimation}
    \label{fig-2b}
  \end{subfigure}
  \begin{subfigure}{0.495\textwidth}
    \centering
    \includegraphics[width=.85\textwidth]{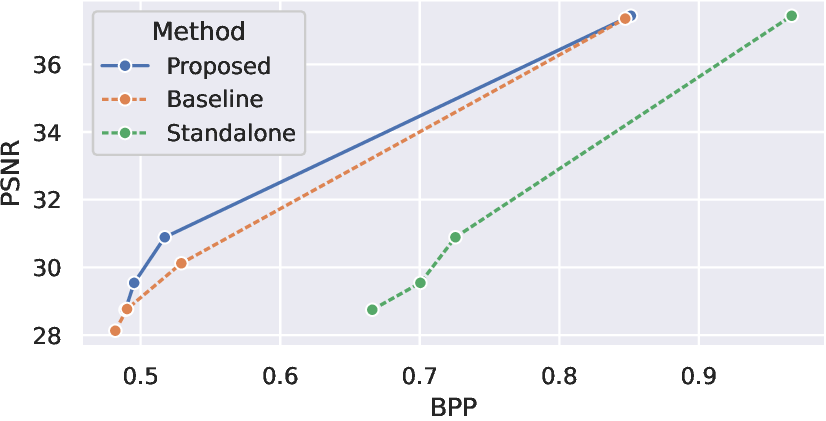}
    \caption{Scalable with object detection}
    \label{fig-2c}
  \end{subfigure}
  \begin{subfigure}{0.495\textwidth}
    \centering
    \includegraphics[width=.85\textwidth]{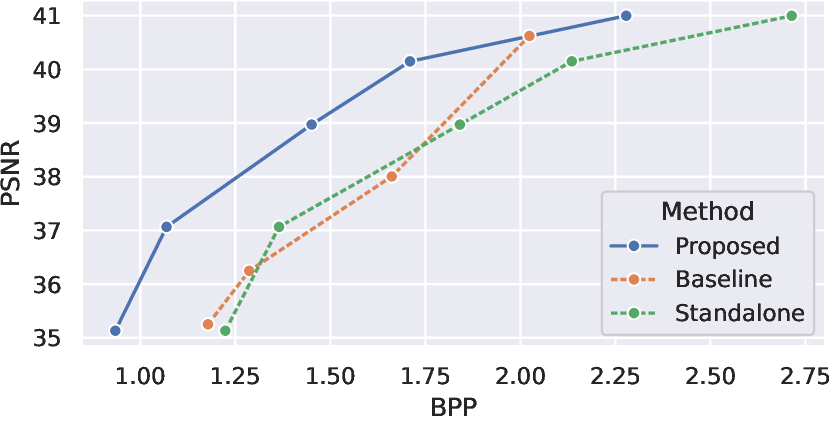}
    \caption{Scalable with depth side-information}
    \label{fig-2d}
  \end{subfigure}
  \begin{subfigure}{0.495\textwidth}
    \centering
    \includegraphics[width=.85\textwidth]{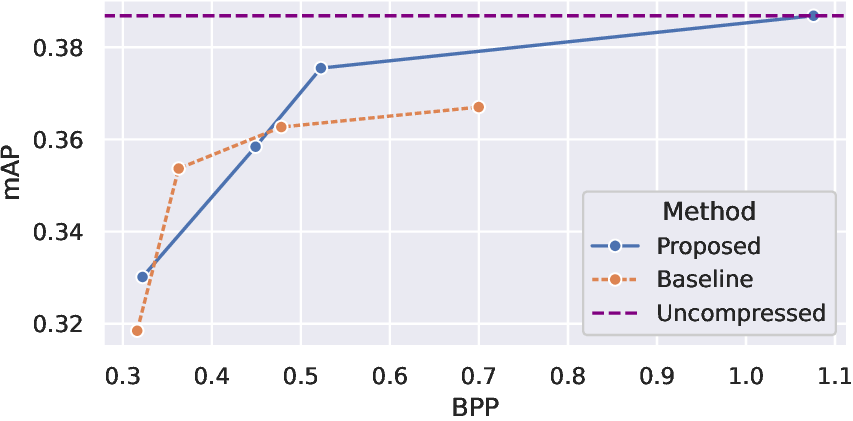}
    \caption{Object detection}
    \label{fig-2e}
  \end{subfigure}
  \begin{subfigure}{0.495\textwidth}
    \centering
    \includegraphics[width=.85\textwidth]{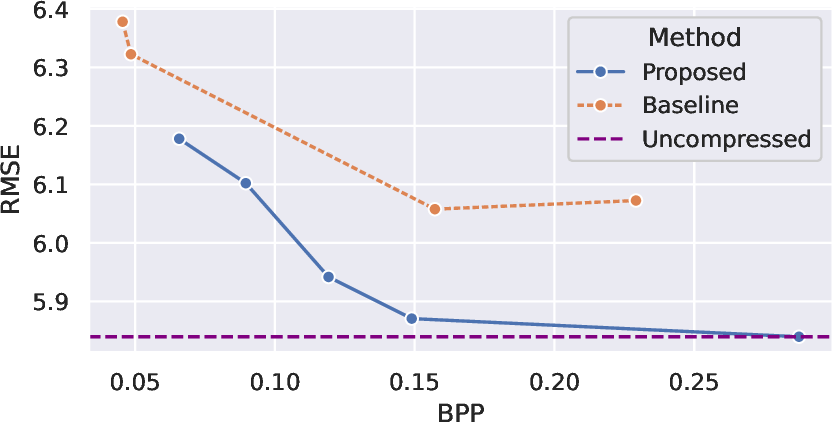}
    \caption{Depth estimation}
    \label{fig-2f}
  \end{subfigure}
  \caption{Rate-distortion performance of base representations on image reconstruction (a)-(d) and base tasks (e)-(f). Left: object detection on COCO 2017; Right: depth estimation on Cityscapes. ``Direct'' means reconstructing the input from the base representation directly, without the dedicated channel (see Fig.~\ref{figure:diagram}). ``Scalable'' means coding reconstruction representation conditional on the base representation; in this case, the BPP reported is the sum of both representations. The ``Standalone'' method does not use side-information. The lines marked ``Uncompressed'' correspond to the best task performance obtained with a very large $\lambda_b = 1,000$, amongst the baseline and proposed methods. PSNR units are in decibels (dB).}
  \label{figure:reconstruction}
\end{figure*}

We perform two different sets of experiments that evaluate the proposed method under different settings. In the first set of experiments, we evaluate the compatibility of compressible learned representations for base tasks against the input reconstruction task. We perform experiments on the COCO 2017 dataset \cite{LinMBHPRDZ14} using object detection as the base task, and on the Cityspaces dataset \cite{cordts2016cityscapes}, using depth estimation as the base task. In the second set of experiments, we evaluate the compatibility of a representation learned for depth estimation as the base task, against semantic segmentation as the secondary task, on Cityscapes. We evaluate the rate-distortion performance of the secondary task in a scalable setting, where we conditionally code the secondary representation using the base representation as side-information. The baseline against which we compare has the same proposed architecture and setup, but with $\beta = 0$ in~(\ref{eq:base_V-info})-(\ref{eq:base_RD}), which is the typical choice for scalable codecs~\cite{Choi2022ScalableIC,ForoutanHAB23}. Hence, the comparison against this baseline assesses whether the additional term with $\beta>0$ improves task compatibility of learned representations. In practice, $\mathbf{g}_a(\cdot; \theta_a)$ is removed from the architecture.

We first train the shared representation on the base task with different values of $\lambda_b$, then we choose a representation from each method producing similar rates and use it to generate side-information for the secondary task. A ``standalone'' method is included in which the base representation is zeroed-out and the entropy model is fine-tuned on a frozen fully-trained analysis transform from the proposed method.

The COCO 2017 dataset consists of 123,287 domain-agnostic images for object detection and segmentation \cite{LinMBHPRDZ14}. We use Faster R-CNN \cite{RenHG017} for object detection with ResNet-50 \cite{HeZRS16} as the back-end. The Cityscapes datasets consists of images of urban scenes. For depth estimation, LRASPP is used with MobileNetV3 as the back-end \cite{HowardPALSCWCTC19}. DeepLabV3 \cite{ChenPSA17} is used for segmentation, with ResNet-50 as back-end.

\subsection{Setup}

We set the dimensionality of the feature space as $M = 192$. Our proposed method uses $\beta = 0.1$. We train using Adam with a fixed learning rate of $1 \times 10^{-4}$, and early-stopping with a patience of 10-50 epochs depending on the task, and a maximum of 1,000 epochs. As data augmentation, we apply random horizontal flips and color jittering. Since the input of Cityscapes has a high resolution, we downscale it by a factor of 2 before processing it and upsample back the resulting predictions before evaluating their performance, ensuring we are still operating on the original problem.

To train the reconstruction tasks, we use the root mean squared error (RMSE) as the distortion function. We compute and report the bits-per-pixel (BPP) using the rate given by the entropy estimates. To speed up the computation of the rate-distortion curve, we train a model for low-compression (highest $\lambda_b$ or $\lambda_e$) and use its parameters as initialization for the models producing the rest of the curve.

\subsection{Input Reconstruction as the Secondary Task}

In this set of experiments, we train representations on the base computer vision tasks and evaluate their performance on image reconstruction. We evaluate the compatibility of these representations in two ways. In the first approach, we train a reconstruction model using the representations generated as immutable input. Since the secondary task is reconstruction, this method highlights how much information regarding the input can be extracted from these representations. In the second approach, we use the representations as side-information to conditionally code a dedicated secondary representation used for reconstruction. This approach shows how compatible the base representation is with the representation trained for image reconstruction.

We generate rate-distortion curves for the 
base task using our approach and the previously described baseline. For each method, we choose one of the models producing a point on the rate-distortion curve as the feature generator for the scalable image reconstruction approach. We choose points with similar BPP values between the two methods.

Fig.~\ref{figure:reconstruction} shows the results obtained. 
For object detection on COCO 2017, the advantage of the proposed method over the baseline is obvious in Fig.~\ref{fig-2a}, but the BD-rate~\cite{bjontegaard2001calculation} could not be computed as the curves do not have an overlap on the vertical axis. The BD-rate advantage over the baseline is 10.3\% in Fig.~\ref{fig-2c} and 1.3\% in Fig.~\ref{fig-2e}, 
showing that the proposed method has a small positive impact on the performance of the base task, while attaining significant gains on the secondary task. For depth estimation on Cityscapes, the proposed method achieves a significant BD-rate gains of 85.2\% and 22.2\% in Figs.~\ref{fig-2b} and~\ref{fig-2d}, respectively, with a BD-Rate gain of 74.8\% in Fig.~\ref{fig-2f}. Again, an advantage on the base task is achieved together with a significant advantage on the secondary task. As discussed previously, this improvement can be explained by the method preventing the co-adaptation to the task and entropy models. 

The image reconstruction performance of the baseline has an interesting behaviour where in low-rate regimes, higher peak signal-to-noise ratio (PSNR) values can be attained. Since the baseline has no reward in creating representations that can be easily used for secondary tasks, the level of compatibility achieved between different representations can vary unexpectedly.

\subsection{Secondary Computer Vision Task}

\begin{figure}
  \centering
  \includegraphics[width=.95\columnwidth]{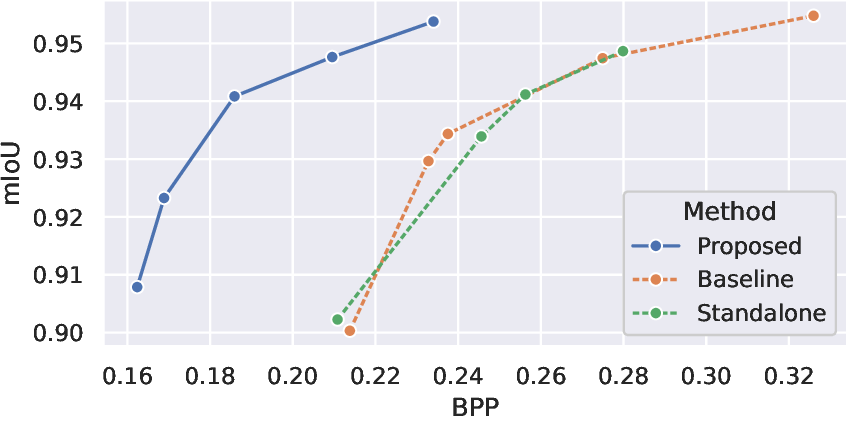}
  \caption{Rate-distortion performance of scalable conditional coding on semantic segmentation on Cityscapes. The BPP measurements corresponds to the sum of the base and secondary rates. Although the distortion function $d_e(\cdot, \cdot)$ used in training is the per-pixel multi-class cross-entropy, we report the more conventional mean Intersection-over-Union (mIoU) test metric.}
  \label{figure:computer-vision}
\end{figure}

To demonstrate that rewarding input reconstruction can also increase the compatibility of the resulting representations with other tasks, we further evaluate the proposed method in a setting where the secondary task is another computer vision task. We employ the same model architecture used earlier on Cityscapes, except that now the secondary task is set to semantic segmentation rather than input reconstruction.  

We run a scalable architecture, in which there is a secondary representation dedicated to semantic segmentation, which is coded conditionally given the base representations used for depth estimation. Fig.~\ref{figure:computer-vision} shows the rate-distortion performance of the proposed method compared to the baseline and ``standalone'' methods on the training set. Due to the small size of the Cityscapes dataset, there is likely some level of overfitting affecting all three methods, which results in fairly high segmentation accuracies in terms of mIoU. Nonetheless, the main comparison here is in terms of the required bitrates, and we see that the proposed method achieves a solid compression gain over the baseline, 26.2\% in terms of the BD-rate. Moreover, the results indicate that the baseline method is unable to extract any useful information from the base representation to support conditional coding of the secondary representation, because it achieves almost the same performance as the ``standalone'' method, which does not use the base representation at all. 

\section{Conclusion}

We studied how well a learned compressible representation for a given task is able to support conditional coding of a secondary task. We demonstrated that in a conventional learned scalable coding, extracting useful information from the base representation is a challenge. However, if the base representation is rewarded for supporting input reconstruction during training, it becomes easier for a conditional codec to use this representation for the secondary task, even if that task is not input reconstruction.  
One possible explanation for this is that, due to the unsupervised nature of the input reconstruction task, the base representation remains more generic and 
useful when extending to multiple, potentially unknown, tasks.

The $\mathcal{V}$-information framework provides a possible interpretation in terms of 
the increase of predictive $\mathcal{V}$-information in the base representation $Y_b$ for a predictive family $\mathcal{V}_e$. 
This would especially be the case when the auxiliary task inducing $Y_b$ is sufficiently non-specific (such as input reconstruction), and $\mathcal{V}_a$ is sufficiently simple, so as to reduce co-adaptation and generalize to other tasks.


\bibliographystyle{IEEEbib}
\bibliography{main}

\end{document}